\documentclass[11pt]{article}

\usepackage[final]{acl}
\usepackage{times}
\usepackage{latexsym}
\usepackage[T1]{fontenc}
\usepackage[utf8]{inputenc}
\usepackage{microtype}
\usepackage{inconsolata}
\usepackage{graphicx}
\usepackage{longtable}
\usepackage{booktabs}

\title{Sample-Size Scaling of the African Languages NLI Evaluation}

\author{
  \textbf{Anuj Tiwari\textsuperscript{1}},
  \textbf{Oluwapelumi Ogunremu\textsuperscript{2}},
  \textbf{Terry Oko-odion\textsuperscript{3}},
  \textbf{Jesujuwon Egbewale\textsuperscript{4}},
  \textbf{Hannah Nwokocha\textsuperscript{5}}
\\
  Noida Institute of Engineering and Technology\textsuperscript{1},
  ML Collective\textsuperscript{1,2,3,4,5}
\\
  \texttt{aj11anuj123@gmail.com},
  \texttt{ ogunremuoluwapelumi@gmail.com},
  \texttt{ terryokoodion@gmail.com},
\\
  \texttt{ egbewalejesujuwon7@gmail.com},
  \texttt{ hannahsopuruchi@gmail.com}
}

\begin{document}
\maketitle
\begin{abstract}
African languages have very little labelled data, and it is unclear if augmenting the quantity of annotation data reliably enhances downstream performance. The study is a systematic sample-size scaling study of natural language inference (NLI) on 16 African languages based on the AfriXNLI benchmark. Under controlled conditions, two multilingual transformer models with roughly 0.6B parameters XLM-R Large fine-tuned on XNLI and AfroXLM-R Large are tested on sample sizes of between 50 and 500 labeled examples and average their results across random subsampling runs. As opposed to the usual belief of monotonic increase with increased data, we find a strongly language-sensitive and often non-monotonic scaling behavior. Some languages show early saturation or decrease in performance with sample size as well as high variance in low resource regimes. These results indicate that the volume of data is not enough to guarantee stable profits to African NLI, creating the necessity of language-sensitive datasets creation and stronger multi-lingual modelling strategies.
\end{abstract}

\section{Introduction}

The latest advancements in NLP have been fuelled by massive pretraining and access to large amounts of labeled data. The advances have however benefited more high resource languages unfairly and many African languages are still underexamined in training and evaluation standards. Therefore, the key problem of multilingual and low-resource NLP is to learn the performance of the models in relation to the amount of labeled data available. One of the most widely used assumptions in machine learning is that as more and more data is labelled, the better the performance will be downstream. Although this assumption is usually true in high-resource environments, it has not yet been carefully studied in the case of the low-resource languages especially the African languages with various typological and morphological characteristics. Practically, annotation is expensive, and expansion of dataset without a clear indication of the benefit can be inefficient or even counterproductive.

This paper provides an analysis of the behavior of natural language inference (NLI) with respect to the amount of labeled data in African language models by using the AfriXNLI benchmark. Rather than proposing new models or datasets, our objective is to empirically characterize scaling behavior, performance stability, and variance across languages and models under controlled experimental conditions. In particular, we answer the following research questions:

\begin{itemize}
    \item \textbf{RQ1}: Does larger labeled data positively affect NLI performance when using African languages in AfriXNLI?
    \item \textbf{RQ2}: How does scaling behavior vary across languages and models?
    \item \textbf{RQ3}: To what extent are observed trends stable under random subsampling?
\end{itemize}

By answering these questions, we hope to offer empirical recommendations to dataset construction and evaluation practice in African NLP, as well as requirements on expectations of data scaling in low-resource semantic reasoning problems.

\section{Related Work}

\textbf{Multilingual and African NLP Benchmarks.}
The current research has contributed to the development of African-language NLP by extending the current standards and developing new assessment tools. AfriXNLI is a human-translated version of the XNLI benchmark of various African languages, allowing to evaluate the human natural language inference in low-resource conditions \citep{afrixnli_dataset} in a unified way.

MasakhaNER offers a named entity recognition system on a large scale on ten African languages, and this project proves that community-driven sets construction are effective in African NLP \citep{adelani2021masakhaner}. AfroLID presents neural language identification toolkit, spanning 517 languages in Africa, and greatly increasing the coverage of languages compared to the previous multi-lingual systems \citep{adebara2022afrolid}. These combined efforts spell out long-term development towards determining assessment materials of the African languages in the context of multilingual NLP.

\textbf{Scaling Data Laws and Efficiency.}
In high-resource settings, language models exhibit predictable scaling behavior. ~\citep{Kaplan2020ScalingLF} demonstrate that language modeling loss reduces according to power-law dependencies on both model size and dataset size. ~\citep{hoffmann2022training} also show that compute-optimal training depends in proportionately more data, as the Chinchilla model can outperform much larger models trained on smaller data. ~\citep{muennighoff2023scaling} however establish that performance improvements reduce quickly in data-constrained environments, and more compute or repeated data produces only small increases in performance. These results cast some doubts on the fact that classical scaling laws can be directly applied to low-resource and multilingual settings.

\textbf{Data Scaling in Low-Resource NLP.}
 \citep{eiselen2023data} look into the impact of the training data size on performance in African languages with particular attention to the morphologically diverse languages of South Africa. They demonstrate that although small data sets can be used to obtain useful models, languages with complex, conjunctive morphology need considerably more data to give similar performance. The importance of linguistic structure in relation to data efficiency is brought out in this work. Nonetheless, they only tested embedding-based models and problems like part-of-speech tagging and they pose the open question of behavior with data scaling in semantic reasoning problems and contemporary fine-tuned pretrained language models.

\textbf{African Languages: Scarcity of Data and Benchmarking.}
Systemic under-representation ~\citep{hussen2025state} report that today only a tiny share of the 2000+ languages of Africa have been trained on modern large language models, and that the field of African languages has been far under-represented 15 compared to its representation across the world. \citep{adebara2024policy} attributes such scarcity to the fact that African languages are structurally unsupported by the current large language model development, and are significantly underrepresented relative to their global distribution. The latest benchmark projects like AfroBench \citep{ojo2024afrobench} and IrokoBench \citep{adelani2024irokobench} extend assessment to African languages and task categories, like reasoning and natural language understanding. Even with this extended coverage, it is evident that these benchmarks always indicate significant performance differences between African and high-resource languages, and that there are still continued issues in modeling and evaluation.

\textbf{Multilingual Representation Models.}
 Multilingual encoders such as mBERT \citep{devlin2019bert} and XLM-R \citep{conneau2020unsupervised}, which have training based on pretraining on multiple languages, are commonly used as multilingual NLP baselines. ~\citep{conneau2020unsupervised} demonstrate that multilingual pretraining significantly improves the cross-lingual test of XLM-R, especially on low-resource languages. Although these models can be shown to be effective in zero-shot transfer, their response to an incremental scale of data of a single African language has not been thoroughly explored. Our study supplements this literature by giving an empirical examination of sample-size scaling action of African-languages NLI.

\section{Experimental Setup}

\subsection{Dataset}

We use the AfriXNLI benchmark, which consists of the sentence pairs of NLI translated to various African languages. We use 16 languages in our experiments which adopt a variety of language families, scripts, and typological properties which are represented in AfriXNLI. All tests are performed on the test splits. Simulating the various labeled data regimes, we adjust the number of test examples to be evaluated by randomly subsampling, but we do not adjust the model parameters.

\subsection{Models}

To allow us to compare pretraining strategies, we assess two multilingual transformer models from similar architectures with about 0.6 billion parameters. The first model, XLM-R Large fine-tuned on XNLI \citep{davison2020xlmrxnli} is a powerful task-aligned multilingual baseline constructed on the XLM-R model \citep{conneau2020unsupervised}. The second model, AfroXLM-R Large \citep{alabi2023afroxlmr} is an African-based form of XLM-R trained with more focus on African-language-based data.

With the choice of similar scale and architecture model, we factor out the influence of pretraining data composition and language coverage and reduce the impact of model size.

\subsection{Evaluation}

For each language–model pair, we evaluate performance at sample sizes ranging from 50 to 500 examples. To control variance due to the selection of the data, we run several random subsampling runs to calculate the mean and standard deviation between the runs of a given sample size.

The most common metric that we report is accuracy, however, we also report precision and F1-score. Such an assessment plan allows us to differentiate between systematic scaling effects and those caused by sampling.

\section{Results}

\subsection{Evaluation Variance between Sample Sizes}

\begin{figure}[t]
    \centering
    \includegraphics[width=\linewidth]{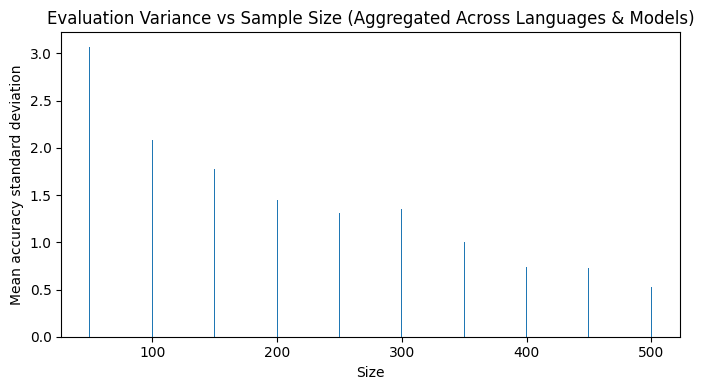}
    \caption{Evaluation variance (standard deviation of accuracy) as a function of sample size, aggregated across all languages and models. The variance decreases very rapidly with the size of the sample meaning that evaluation regimes with low resources are unstable.}
    \label{fig:var_all}
\end{figure}

We initially analyse the patterns of the evaluation variance with the change in size of the sample. The standard deviation of the accuracy is reported in Figure~\ref{fig:var_all} in the aggregate form over all the languages and models. The maximum variance is in low-resource samples (50-100 examples) and is sharply decreasing with increase in the sample size that reaches a point of an average of 300 samples after which the variance is constant.

This tendency shows that small sets of evaluation produce very unstable performance estimates which are highly dependent on the specific sets of samples that one is analyzing. With an increase in sample size, the variance decreases implying that the bigger the evaluation sets, the more true model performance is likely to have been estimated.

\begin{figure}[t]
    \centering
    \includegraphics[width=\linewidth]{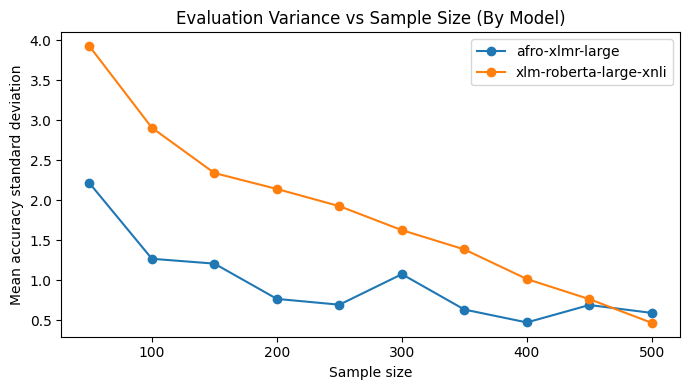}
    \caption{Comparison of evaluation variance with sample size, separately by XLM-R Large and AfroXLM-R Large. Both models show high variance in low-resource regimes and stabilize with larger evaluation sets.}
    \label{fig:var_model}
\end{figure}

Figure~\ref{fig:var_model} also further modifies this effect by model. Both XLM-R Large and AfroXLM-R Large have the following qualitative trend: large variance at the beginning of the sample size, and this is followed by a sharp rise in the value of the sample size. Although the absolute levels of different variance are slightly different, the overall trend is maintained in all models which implies that the instability of evaluation is not particular to a given pretraining strategy.

\subsection{Trends of Global Scaling across Languages and Models}

\begin{figure}[t!]
    \centering
    \includegraphics[width=0.85\linewidth]{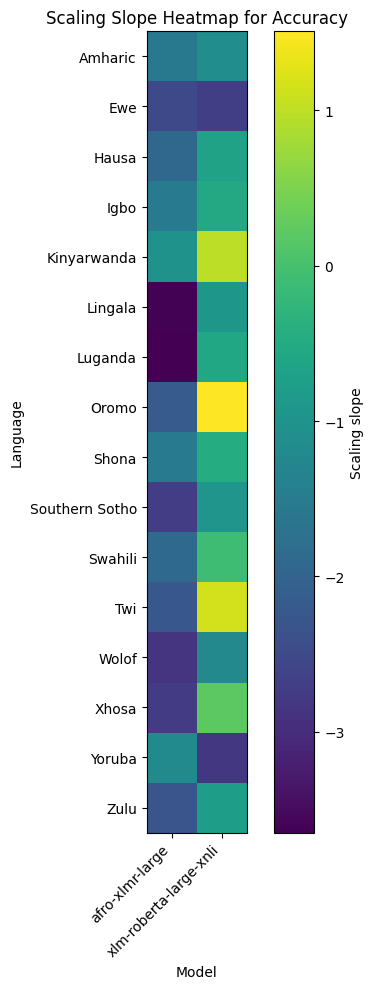}
    \caption{Scaling slope heatmap on accuracy between language-model pairs. Positive slopes mean that the performance improves as the sample size increases and negative slope means that the performance deteriorates.}
    \label{fig:slope_heatmap}
\end{figure}

As the sample size increases, the variance reduces, however, accuracy may not necessarily increase monotonically. Figure~\ref{fig:slope_heatmap} provides a heatmap of scaling slopes of every language model pair, indicating whether the performance improves, stays constant or reduces with increased sample size.

The heatmap indicates that there is significant heterogeneity in languages. There are some nearly zero or slightly positive slopes, that is, weak gains or early saturation, and also negative slopes, that is, systematic degradation in performance with increasing evaluation sets becoming larger and more representative. Such patterns exist in both models, indicating that the scaling behaviour is highly language-specific, as opposed to being model-driven.

\subsection{Scaling Behavior Dependent on Language: Yoruba vs Kinyarwanda}

\begin{figure}[t]
    \centering
    \includegraphics[width=\linewidth]{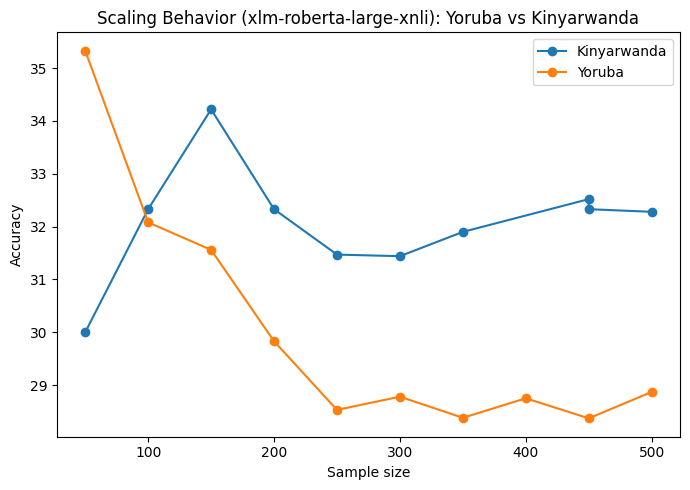}
    \caption{Yoruba and Kinyarwanda evaluation scaling behaviour with XLM-R Large. Yoruba experiences monotonic deterioration as the sample size increases and Kinyarwanda experiences initial improvement and afterwards saturation.}
    \label{fig:yoruba_kinya_xlmr}
\end{figure}

To illustrate these trends concretely, we analyze scaling behavior for Yoruba and Kinyarwanda under each model. Figure~\ref{fig:yoruba_kinya_xlmr} shows results for XLM-R Large. Yoruba exhibits pronounced small-sample optimism, with relatively high accuracy at 50 examples followed by a consistent decline as sample size increases. This monotonic degradation suggests that small evaluation subsets overestimate performance, masking systematic errors that emerge with broader coverage.

Conversely, there is a slight rise in performance of Kinyarwanda up to around 150 examples after which it starts to decrease and stabilize. At bigger sample sizes, variance collapses, meaning that it can no longer be measured with its actual performance level by the smaller subsets.

\begin{figure}[t]
    \centering
    \includegraphics[width=\linewidth]{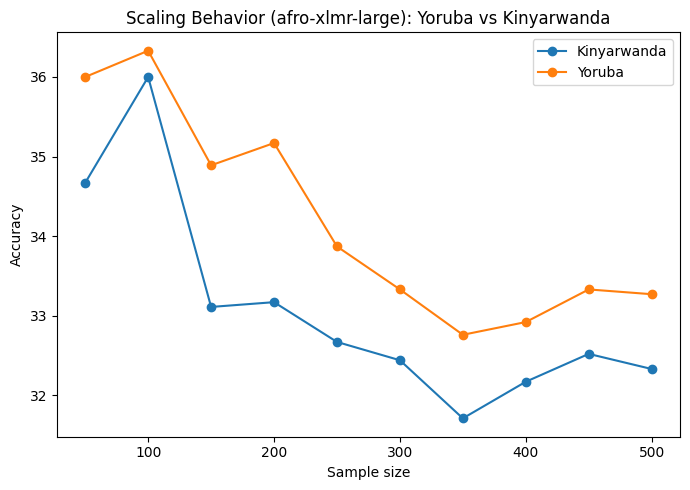}
    \caption{Yoruba and Kinyarwanda AfroXLM-R Large evaluation scaling behavior. The non-monotonic tendencies that are specific to language prevail within models.}
    \label{fig:yoruba_kinya_afro}
\end{figure}

The same comparison is made in Figure~\ref{fig:yoruba_kinya_afro} on AfroXLM-R Large. The qualitative trends are similar: Yoruba demonstrates again decreasing accuracy with the growing sample size at the same time Kinyarwanda demonstrates initial gains and then stabilities. The fact that such trends are maintained in the models supports the conclusion that the scaling behavior is more language-specific than model-specific.

\section{Discussion}

Our findings dispute the widely held belief that more uniformly augmented data is always better to use when applying African-language natural language inference. In languages, non-monotonic scaling behavior in evaluation accuracy is observed, where information other than data quantity influences it, e.g. the distribution of labels, ambiguity in translations and representativeness of evaluation subsets contribute significantly.

Notably, we come up with conclusions about AfriXNLI only and these need not be generalized as being applicable in all African languages. However, the similarity in trends observed at random subsampling runs and between two different multilingual models indicates that it is systematic, and not incidental.

\paragraph{Evaluation bias and small-sample optimism}
Among the main conclusions that this research made is that small sample evaluation is systematically biased in overestimating the performance of the models. The figure below (Figures~1 and~2) indicated that at sample sizes smaller than 200 evaluation variance is high resulting in unstable and optimistic bias accuracy estimates. With increase in sample size, both variance and accuracy collapse and this is not due to decreasing model quality but rather due to harder instances, neutral cases, and translation ambiguities displayed by larger evaluation sets that are underrepresented in small samples. This point is crucial: we are examining reliability of evaluation and not learning curves, and the apparent decline in performance indicates the lower estimation bias and not the alterations of the model behavior.

\paragraph{Language-specific non-monotonic scaling}
Figure~3, the scaling slope heatmap, indicates that there is a lot of heterogeneity between languages. Although in some languages, the slopes are weakly positive or near-zero, in others, the slopes are negative, which means that the performance of a language deteriorates as sets of evaluations increase. These patterns are present in both models and it is revealed that scaling behavior is more language dependent than model dependent. The differences between Yoruba and Kinyarwanda case studies, Figures~4 and~5, illustrate this contrast quite well: Yoruba has high small-sample optimism and decreases in a monotonic way, whereas Kinyarwanda has small gains at the start and stagnates. These variations indicate that an individual set evaluation size might provide inaccurate results when used across languages when used consistently.

\paragraph{Pretraining strategies Model effects}
Even though AfroXLM-R Large can achieve significantly better accuracy and reduced variance at very small sample sizes than XLM-R Large fine-tuned on XNLI, both models have similar qualitative scaling behaviors across languages. Africa-centric pretraining improves initial stability but does not eliminate non-monotonic scaling or language specific evaluation bias. This implies that data composition pretraining is not enough to consider the heterogeneity of African-language NLI assessment and that the choice of the model cannot influence not only stability in the evaluation but also absolute performance.

\paragraph{Benchmarking and evaluation implications}
These findings have direct methodological implications. Single scores on small test sets of accuracy can significantly exaggerate the ability of models to perform under low resource conditions. The larger sets of evaluation decrease variance and bias but can indicate lower real performance and therefore makes comparing studies more difficult. We also suggest that African language benchmarks should report variance between subsamples, should not over rely on small held-out sets, and should take into account language specific evaluation sizes instead of fixed-size test sets.

Overall, we find that the amount of data is not sufficient to ensure credible assessment on the part of African NLI. Rather, meaningful benchmarking in a low resource multilingual setting requires representative sampling, meticulous dataset construction, and stability analysis to contribute to it.

\paragraph{Evaluation stability and saturation}
In order to measure stability in evaluation, we approximate a saturation value of each language-model combination, which is the minimum size of the evaluation where the mean accuracy is varied by at most $\pm$0.5\%. The smallest sample size at which additional increments in evaluation data do not produce significant performance variations is called the $n^{*}$.
\[
n^{*} = \min \left\{ n \;\middle|\; \max_{m > n} \left| A(m) - A(n) \right| \le \epsilon \right\}.
\]
In case there is no $n \le 500$, we declare saturation point as $>500$.
We are not training, but only testing. Here saturation refers to the amount of evaluation data one only needs to achieve the stabilization of the estimated performance.

Table~\ref{tab:saturation} provides the summary of these saturation points in all languages and models. We find very great language to language disparity. In some languages (e.g., Shona, Yoruba, Xhosa) as few as 200 or 250 evaluation samples can give the required stable performance estimates, whereas in others (e.g., Swahili, Kinyarwanda, Oromo) it can take 400 or 450 samples. It is worth noting that Wolof underestimates performance even in 500 samples of XLM-R Large, and this implies that Wolof remains unstable in performance estimation. 

These variations are much the same across models indicating that language and dataset specific factors are the primary causes of saturation behaviour as opposed to model architecture itself. Overall, these findings indicate that model performance can be significantly misestimated using fixed-size evaluation benchmarks in the case of African languages, and that the required volume of evaluation data to make reliable estimates can differ significantly across languages.

\begin{table}[t]
\centering
\small
\caption{The saturation points are estimated at a sample size of ">500" at which the average error levels off within $\pm$0.5\% between the languages and models.}
\label{tab:saturation}
\begin{tabular}{lcc}
\toprule
Language & xlm-roberta-large-xnli & afro-xlmr-large \\
\midrule
Amharic        & 450  & 250 \\
Ewe            & 250  & 400 \\
Hausa          & 400  & 400 \\
Igbo           & 300  & 250 \\
Kinyarwanda    & 450  & 400 \\
Lingala        & 300  & 350 \\
Luganda        & 350  & 450 \\
Oromo          & 400  & 400 \\
Shona          & 200  & 400 \\
Southern Sotho & 300  & 350 \\
Swahili        & 300  & 450 \\
Twi            & 450  & 400 \\
Wolof          & >500 & 450 \\
Xhosa          & 250  & 400 \\
Yoruba         & 250  & 400 \\
Zulu           & 300  & 400 \\
\bottomrule
\end{tabular}
\end{table}

\section{Limitations}

The limitations of our study are as follows:

\begin{itemize}
    \item \textbf{Dataset scope} All experiments are conducted on AfriXNLI; thus, observed trends may reflect dataset-specific properties such as translation artifacts or label distribution biases.
    
    \item \textbf{Evaluation vs learning} We get the evaluation behavior as opposed to the dynamics of learning. Models do not optimize on successively large training sets hence results reflect the stability and bias of the performance estimates, not the improvement in performance with more training data.
    
    \item \textbf{Model scale} The scale of experiments is restricted to 2 models of multilingual size 0.6b. Relationships between data scaling and scales of models are not investigated.
\end{itemize}

Regardless of these constraints, the fact that trends were similar among languages, models, and random subsampling runs implies that we have been able to capture systematic elements of evaluation reliability to African NLI.

\section{Future Work}

This analysis can be developed in several ways in future work:

\begin{itemize}
    \item \textbf{Broader tasks and datasets} The generalizability of the found evaluated scaling behaviour of the study should be tested by extending the study to other African NLP benchmark tasks, like sentiment analysis or named entity recognition.
    
    \item \textbf{Linguistic and dataset effects} Adding linguistic metadata, label distributions and tokenization statistics can be useful to explain language behavioral specifics of saturation and non-monotonic scaling.
    
    \item \textbf{Learning dynamics} Learning scaling behavior: Fine-tuning, but not evaluation, would help illuminate the effect of the addition of labeled data on real model learning with African languages.
\end{itemize}

\section{Conclusion}

Here we provide a detailed study of scale behavior in terms of sample-size on the AfriXNLI benchmark using African languages. Our results based on controlled evaluation in 16 languages and two multilingual models and a series of random subsampling runs demonstrate that growth in evaluation data does not come at uniform or monotonic benefits. Rather, scaling behavior is very language-specific, usually non monotonic, and hugely influenced by evaluation variation under low resource limitations.

We show that the performance estimates of small evaluation subsets are often optimistically biased, whereas the estimates of larger subsets are indicative of latent difficulty and their estimates are more stable. This demonstrates the difference between the evaluation reliability and model learning as one of the key issues in the African NLP.

On our results, we suggest: (i) do not use single-point assessment on very small test sets and report the mean $\pm$ standard deviation on more than one subsample, (ii) supplement aggregate measures with per-class measures and (iii) use at least 300 evaluation samples unless otherwise, and consider results below this scale as high-noise measures. In a broader sense, our paper warns on naive beliefs about the reliability of increased information as a means to have dependable evaluation and the need to have practices that are evaluation conscious in benchmarking of African languages.

\bibliography{references}

\appendix

\section{Appendix - Full Results for models}

\begin{table}[t]
\centering
\small
\caption{Evaluation of Swahili with \texttt{xlm-roberta-large-xnli}. Measures are provided of means $\pm$ standard deviation.}
\label{tab:xlmr-swahili}
\begin{tabular}{rccc}
\toprule
Size & Accuracy (\%) & Precision & F1 \\
\midrule
50  & 33.33 $\pm$ 1.89 & 0.339 $\pm$ 0.009 & 0.331 $\pm$ 0.014 \\
100 & 34.08 $\pm$ 1.41 & 0.341 $\pm$ 0.014 & 0.340 $\pm$ 0.014 \\
150 & 32.89 $\pm$ 0.31 & 0.332 $\pm$ 0.002 & 0.330 $\pm$ 0.008 \\
200 & 33.33 $\pm$ 0.85 & 0.335 $\pm$ 0.004 & 0.334 $\pm$ 0.006 \\
250 & 33.87 $\pm$ 1.61 & 0.341 $\pm$ 0.015 & 0.340 $\pm$ 0.016 \\
300 & 32.89 $\pm$ 1.59 & 0.331 $\pm$ 0.015 & 0.330 $\pm$ 0.015 \\
350 & 32.95 $\pm$ 1.52 & 0.332 $\pm$ 0.015 & 0.331 $\pm$ 0.015 \\
400 & 33.25 $\pm$ 1.14 & 0.335 $\pm$ 0.011 & 0.334 $\pm$ 0.011 \\
450 & 33.19 $\pm$ 0.93 & 0.333 $\pm$ 0.010 & 0.333 $\pm$ 0.009 \\
500 & 33.13 $\pm$ 0.41 & 0.331 $\pm$ 0.004 & 0.331 $\pm$ 0.004 \\
\bottomrule
\end{tabular}
\end{table}

\begin{table}[t]
\centering
\small
\caption{Evaluation of Lingala with \texttt{xlm-roberta-large-xnli}. Measures are provided of means $\pm$ standard deviation.}
\label{tab:xlmr-lingala}
\begin{tabular}{rccc}
\toprule
Size & Accuracy (\%) & Precision & F1 \\
\midrule
50  & 34.00 $\pm$ 0.00 & 0.346 $\pm$ 0.027 & 0.323 $\pm$ 0.010 \\
100 & 29.67 $\pm$ 1.70 & 0.293 $\pm$ 0.026 & 0.281 $\pm$ 0.021 \\
150 & 31.33 $\pm$ 0.94 & 0.308 $\pm$ 0.028 & 0.294 $\pm$ 0.014 \\
200 & 30.83 $\pm$ 1.84 & 0.301 $\pm$ 0.030 & 0.284 $\pm$ 0.022 \\
250 & 30.67 $\pm$ 1.91 & 0.302 $\pm$ 0.029 & 0.283 $\pm$ 0.022 \\
300 & 31.89 $\pm$ 1.66 & 0.314 $\pm$ 0.018 & 0.296 $\pm$ 0.018 \\
350 & 31.71 $\pm$ 1.46 & 0.313 $\pm$ 0.015 & 0.294 $\pm$ 0.015 \\
400 & 31.67 $\pm$ 1.36 & 0.309 $\pm$ 0.015 & 0.293 $\pm$ 0.014 \\
450 & 31.93 $\pm$ 0.90 & 0.311 $\pm$ 0.008 & 0.295 $\pm$ 0.008 \\
500 & 31.87 $\pm$ 0.82 & 0.312 $\pm$ 0.008 & 0.294 $\pm$ 0.008 \\
\bottomrule
\end{tabular}
\end{table}

\begin{table}[t]
\centering
\small
\caption{Evaluation of Igbo with \texttt{xlm-roberta-large-xnli}. Measures are provided of means $\pm$ standard deviation.}
\label{tab:xlmr-igbo}
\begin{tabular}{rccc}
\toprule
Size & Accuracy (\%) & Precision & F1 \\
\midrule
50  & 31.33 $\pm$ 1.89 & 0.424 $\pm$ 0.008 & 0.314 $\pm$ 0.026 \\
100 & 32.00 $\pm$ 3.56 & 0.456 $\pm$ 0.011 & 0.314 $\pm$ 0.040 \\
150 & 32.44 $\pm$ 2.06 & 0.458 $\pm$ 0.036 & 0.315 $\pm$ 0.024 \\
200 & 31.33 $\pm$ 1.93 & 0.441 $\pm$ 0.020 & 0.306 $\pm$ 0.020 \\
250 & 30.67 $\pm$ 1.80 & 0.424 $\pm$ 0.009 & 0.298 $\pm$ 0.019 \\
300 & 29.89 $\pm$ 1.50 & 0.406 $\pm$ 0.015 & 0.299 $\pm$ 0.018 \\
350 & 29.62 $\pm$ 1.28 & 0.408 $\pm$ 0.013 & 0.299 $\pm$ 0.014 \\
400 & 30.08 $\pm$ 0.82 & 0.403 $\pm$ 0.009 & 0.296 $\pm$ 0.010 \\
450 & 29.85 $\pm$ 0.46 & 0.406 $\pm$ 0.005 & 0.293 $\pm$ 0.005 \\
500 & 30.08 $\pm$ 0.33 & 0.409 $\pm$ 0.005 & 0.294 $\pm$ 0.003 \\
\bottomrule
\end{tabular}
\end{table}

\begin{table}[t]
\centering
\small
\caption{Evaluation of Hausa with \texttt{xlm-roberta-large-xnli}. Measures are provided of means $\pm$ standard deviation.}
\label{tab:xlmr-hausa}
\begin{tabular}{rccc}
\toprule
Size & Accuracy (\%) & Precision & F1 \\
\midrule
50  & 32.67 $\pm$ 2.49 & 0.317 $\pm$ 0.032 & 0.316 $\pm$ 0.026 \\
100 & 32.67 $\pm$ 1.79 & 0.335 $\pm$ 0.007 & 0.327 $\pm$ 0.011 \\
150 & 31.78 $\pm$ 1.37 & 0.317 $\pm$ 0.017 & 0.316 $\pm$ 0.015 \\
200 & 32.80 $\pm$ 1.08 & 0.314 $\pm$ 0.006 & 0.316 $\pm$ 0.008 \\
250 & 32.53 $\pm$ 2.93 & 0.325 $\pm$ 0.026 & 0.325 $\pm$ 0.027 \\
300 & 32.56 $\pm$ 3.22 & 0.326 $\pm$ 0.029 & 0.325 $\pm$ 0.031 \\
350 & 32.10 $\pm$ 1.94 & 0.321 $\pm$ 0.021 & 0.321 $\pm$ 0.020 \\
400 & 31.42 $\pm$ 1.84 & 0.317 $\pm$ 0.016 & 0.315 $\pm$ 0.017 \\
450 & 31.19 $\pm$ 1.18 & 0.317 $\pm$ 0.012 & 0.314 $\pm$ 0.012 \\
500 & 31.13 $\pm$ 0.38 & 0.318 $\pm$ 0.006 & 0.314 $\pm$ 0.005 \\
\bottomrule
\end{tabular}
\end{table}

\begin{table}[t]
\centering
\small
\caption{Evaluation of Yoruba with \texttt{xlm-roberta-large-xnli}. Measures are provided of means $\pm$ standard deviation.}
\label{tab:xlmr-yoruba}
\begin{tabular}{rccc}
\toprule
Size & Accuracy (\%) & Precision & F1 \\
\midrule
50  & 35.33 $\pm$ 2.49 & 0.439 $\pm$ 0.060 & 0.342 $\pm$ 0.035 \\
100 & 32.08 $\pm$ 2.83 & 0.421 $\pm$ 0.077 & 0.296 $\pm$ 0.026 \\
150 & 31.56 $\pm$ 2.27 & 0.396 $\pm$ 0.046 & 0.291 $\pm$ 0.020 \\
200 & 29.83 $\pm$ 1.43 & 0.391 $\pm$ 0.020 & 0.283 $\pm$ 0.021 \\
250 & 28.53 $\pm$ 1.24 & 0.367 $\pm$ 0.010 & 0.268 $\pm$ 0.020 \\
300 & 28.78 $\pm$ 0.68 & 0.359 $\pm$ 0.007 & 0.271 $\pm$ 0.013 \\
350 & 28.38 $\pm$ 0.36 & 0.353 $\pm$ 0.005 & 0.268 $\pm$ 0.008 \\
400 & 28.75 $\pm$ 0.89 & 0.369 $\pm$ 0.017 & 0.272 $\pm$ 0.011 \\
450 & 28.37 $\pm$ 1.06 & 0.358 $\pm$ 0.018 & 0.269 $\pm$ 0.011 \\
500 & 28.87 $\pm$ 0.52 & 0.358 $\pm$ 0.011 & 0.273 $\pm$ 0.005 \\
\bottomrule
\end{tabular}
\end{table}

\begin{table}[t]
\centering
\small
\caption{Evaluation of Kinyarwanda with \texttt{xlm-roberta-large-xnli}. Measures are provided of means $\pm$ standard deviation.}
\label{tab:xlmr-kinyarwanda}
\begin{tabular}{rccc}
\toprule
Size & Accuracy (\%) & Precision & F1 \\
\midrule
50  & 30.00 $\pm$ 4.32 & 0.455 $\pm$ 0.064 & 0.305 $\pm$ 0.031 \\
100 & 32.33 $\pm$ 3.09 & 0.430 $\pm$ 0.040 & 0.325 $\pm$ 0.023 \\
150 & 34.22 $\pm$ 5.45 & 0.440 $\pm$ 0.040 & 0.337 $\pm$ 0.050 \\
200 & 32.33 $\pm$ 4.50 & 0.420 $\pm$ 0.041 & 0.318 $\pm$ 0.043 \\
250 & 31.47 $\pm$ 3.09 & 0.412 $\pm$ 0.019 & 0.311 $\pm$ 0.030 \\
300 & 31.44 $\pm$ 1.75 & 0.402 $\pm$ 0.012 & 0.309 $\pm$ 0.018 \\
350 & 31.90 $\pm$ 1.75 & 0.406 $\pm$ 0.015 & 0.312 $\pm$ 0.016 \\
400 & 32.33 $\pm$ 1.45 & 0.415 $\pm$ 0.011 & 0.319 $\pm$ 0.015 \\
450 & 32.52 $\pm$ 0.38 & 0.412 $\pm$ 0.004 & 0.320 $\pm$ 0.003 \\
500 & 32.28 $\pm$ 0.16 & 0.403 $\pm$ 0.001 & 0.315 $\pm$ 0.003 \\
\bottomrule
\end{tabular}
\end{table}

\begin{table}[t]
\centering
\small
\caption{Evaluation of Zulu with \texttt{xlm-roberta-large-xnli}. Measures are provided of means $\pm$ standard deviation.}
\label{tab:xlmr-zulu}
\begin{tabular}{rccc}
\toprule
Size & Accuracy (\%) & Precision & F1 \\
\midrule
50  & 32.08 $\pm$ 4.32 & 0.337 $\pm$ 0.057 & 0.319 $\pm$ 0.048 \\
100 & 31.33 $\pm$ 2.49 & 0.349 $\pm$ 0.026 & 0.323 $\pm$ 0.029 \\
150 & 33.33 $\pm$ 0.54 & 0.358 $\pm$ 0.011 & 0.342 $\pm$ 0.009 \\
200 & 31.17 $\pm$ 2.09 & 0.359 $\pm$ 0.012 & 0.327 $\pm$ 0.016 \\
250 & 30.88 $\pm$ 1.96 & 0.351 $\pm$ 0.014 & 0.326 $\pm$ 0.017 \\
300 & 30.56 $\pm$ 2.01 & 0.341 $\pm$ 0.012 & 0.328 $\pm$ 0.017 \\
350 & 30.67 $\pm$ 2.38 & 0.340 $\pm$ 0.018 & 0.321 $\pm$ 0.021 \\
400 & 30.80 $\pm$ 1.47 & 0.338 $\pm$ 0.010 & 0.316 $\pm$ 0.013 \\
450 & 30.07 $\pm$ 0.93 & 0.339 $\pm$ 0.006 & 0.317 $\pm$ 0.008 \\
500 & 30.33 $\pm$ 0.77 & 0.341 $\pm$ 0.004 & 0.319 $\pm$ 0.006 \\
\bottomrule
\end{tabular}
\end{table}

\begin{table}[t]
\centering
\small
\caption{Evaluation of Amharic with \texttt{xlm-roberta-large-xnli}. Measures are provided of means $\pm$ standard deviation.}
\label{tab:xlmr-amharic}
\begin{tabular}{rccc}
\toprule
Size & Accuracy (\%) & Precision & F1 \\
\midrule
50  & 33.33 $\pm$ 6.18 & 0.333 $\pm$ 0.058 & 0.332 $\pm$ 0.059 \\
100 & 33.67 $\pm$ 4.11 & 0.333 $\pm$ 0.043 & 0.334 $\pm$ 0.041 \\
150 & 32.08 $\pm$ 3.49 & 0.332 $\pm$ 0.022 & 0.325 $\pm$ 0.028 \\
200 & 31.58 $\pm$ 2.27 & 0.333 $\pm$ 0.020 & 0.323 $\pm$ 0.021 \\
250 & 32.13 $\pm$ 1.36 & 0.337 $\pm$ 0.086 & 0.329 $\pm$ 0.010 \\
300 & 31.67 $\pm$ 0.82 & 0.327 $\pm$ 0.094 & 0.321 $\pm$ 0.006 \\
350 & 31.85 $\pm$ 0.59 & 0.327 $\pm$ 0.092 & 0.318 $\pm$ 0.006 \\
400 & 31.88 $\pm$ 0.82 & 0.329 $\pm$ 0.009 & 0.319 $\pm$ 0.008 \\
450 & 30.96 $\pm$ 0.91 & 0.329 $\pm$ 0.010 & 0.319 $\pm$ 0.010 \\
500 & 30.73 $\pm$ 0.52 & 0.327 $\pm$ 0.006 & 0.317 $\pm$ 0.004 \\
\bottomrule
\end{tabular}
\end{table}

\begin{table}[t]
\centering
\small
\caption{Evaluation of Southern sotho with \texttt{xlm-roberta-large-xnli}. Measures are provided of means $\pm$ standard deviation.}
\label{tab:xlmr-southern-sotho}
\begin{tabular}{rccc}
\toprule
Size & Accuracy (\%) & Precision & F1 \\
\midrule
50  & 32.00 $\pm$ 3.27 & 0.420 $\pm$ 0.040 & 0.308 $\pm$ 0.036 \\
100 & 28.33 $\pm$ 4.11 & 0.397 $\pm$ 0.022 & 0.268 $\pm$ 0.027 \\
150 & 29.78 $\pm$ 4.43 & 0.383 $\pm$ 0.093 & 0.278 $\pm$ 0.036 \\
200 & 30.08 $\pm$ 4.14 & 0.399 $\pm$ 0.027 & 0.285 $\pm$ 0.037 \\
250 & 29.73 $\pm$ 2.54 & 0.376 $\pm$ 0.021 & 0.281 $\pm$ 0.023 \\
300 & 29.44 $\pm$ 2.20 & 0.355 $\pm$ 0.028 & 0.277 $\pm$ 0.021 \\
350 & 29.14 $\pm$ 1.42 & 0.354 $\pm$ 0.028 & 0.275 $\pm$ 0.013 \\
400 & 29.33 $\pm$ 1.25 & 0.357 $\pm$ 0.020 & 0.279 $\pm$ 0.011 \\
450 & 29.48 $\pm$ 0.46 & 0.356 $\pm$ 0.011 & 0.280 $\pm$ 0.004 \\
500 & 29.80 $\pm$ 0.49 & 0.357 $\pm$ 0.010 & 0.282 $\pm$ 0.004 \\
\bottomrule
\end{tabular}
\end{table}

\begin{table}[t]
\centering
\small
\caption{Evaluation of Oromo with \texttt{xlm-roberta-large-xnli}. Measures are provided of means $\pm$ standard deviation.}
\label{tab:xlmr-oromo}
\begin{tabular}{rccc}
\toprule
Size & Accuracy (\%) & Precision & F1 \\
\midrule
50  & 24.00 $\pm$ 12.96 & 0.269 $\pm$ 0.118 & 0.246 $\pm$ 0.126 \\
100 & 27.67 $\pm$ 4.64  & 0.333 $\pm$ 0.047 & 0.295 $\pm$ 0.046 \\
150 & 28.44 $\pm$ 2.57  & 0.347 $\pm$ 0.022 & 0.305 $\pm$ 0.023 \\
200 & 28.00 $\pm$ 2.04  & 0.341 $\pm$ 0.026 & 0.300 $\pm$ 0.022 \\
250 & 28.80 $\pm$ 1.31  & 0.346 $\pm$ 0.011 & 0.308 $\pm$ 0.013 \\
300 & 28.89 $\pm$ 0.83  & 0.342 $\pm$ 0.019 & 0.306 $\pm$ 0.011 \\
350 & 28.76 $\pm$ 0.97  & 0.342 $\pm$ 0.019 & 0.306 $\pm$ 0.012 \\
400 & 27.67 $\pm$ 0.82  & 0.337 $\pm$ 0.013 & 0.297 $\pm$ 0.010 \\
450 & 28.15 $\pm$ 0.90  & 0.339 $\pm$ 0.011 & 0.301 $\pm$ 0.010 \\
500 & 27.47 $\pm$ 0.34  & 0.333 $\pm$ 0.005 & 0.294 $\pm$ 0.004 \\
\bottomrule
\end{tabular}
\end{table}

\begin{table}[t]
\centering
\small
\caption{Evaluation of Twi with \texttt{xlm-roberta-large-xnli}. Measures are provided of means $\pm$ standard deviation.}
\label{tab:xlmr-twi}
\begin{tabular}{rccc}
\toprule
Size & Accuracy (\%) & Precision & F1 \\
\midrule
50  & 25.33 $\pm$ 4.71 & 0.276 $\pm$ 0.034 & 0.214 $\pm$ 0.034 \\
100 & 27.00 $\pm$ 3.74 & 0.319 $\pm$ 0.064 & 0.225 $\pm$ 0.033 \\
150 & 27.78 $\pm$ 3.94 & 0.349 $\pm$ 0.049 & 0.237 $\pm$ 0.040 \\
200 & 27.00 $\pm$ 2.83 & 0.332 $\pm$ 0.053 & 0.231 $\pm$ 0.029 \\
250 & 26.00 $\pm$ 2.47 & 0.316 $\pm$ 0.045 & 0.224 $\pm$ 0.023 \\
300 & 26.78 $\pm$ 1.50 & 0.302 $\pm$ 0.031 & 0.232 $\pm$ 0.015 \\
350 & 27.33 $\pm$ 1.66 & 0.322 $\pm$ 0.017 & 0.240 $\pm$ 0.015 \\
400 & 27.08 $\pm$ 1.01 & 0.324 $\pm$ 0.022 & 0.237 $\pm$ 0.008 \\
450 & 27.93 $\pm$ 0.21 & 0.334 $\pm$ 0.003 & 0.244 $\pm$ 0.001 \\
500 & 28.00 $\pm$ 0.43 & 0.328 $\pm$ 0.014 & 0.245 $\pm$ 0.005 \\
\bottomrule
\end{tabular}
\end{table}

\begin{table}[t]
\centering
\small
\caption{Evaluation of Shona with \texttt{xlm-roberta-large-xnli}. Measures are provided of means $\pm$ standard deviation.}
\label{tab:xlmr-shona}
\begin{tabular}{rccc}
\toprule
Size & Accuracy (\%) & Precision & F1 \\
\midrule
50  & 26.00 $\pm$ 5.66 & 0.301 $\pm$ 0.122 & 0.240 $\pm$ 0.064 \\
100 & 23.00 $\pm$ 4.55 & 0.256 $\pm$ 0.104 & 0.208 $\pm$ 0.050 \\
150 & 25.33 $\pm$ 3.81 & 0.319 $\pm$ 0.063 & 0.237 $\pm$ 0.043 \\
200 & 24.50 $\pm$ 2.27 & 0.315 $\pm$ 0.055 & 0.233 $\pm$ 0.029 \\
250 & 24.53 $\pm$ 2.00 & 0.311 $\pm$ 0.032 & 0.233 $\pm$ 0.024 \\
300 & 24.67 $\pm$ 1.52 & 0.307 $\pm$ 0.027 & 0.234 $\pm$ 0.018 \\
350 & 24.18 $\pm$ 0.94 & 0.298 $\pm$ 0.013 & 0.228 $\pm$ 0.011 \\
400 & 24.42 $\pm$ 1.30 & 0.314 $\pm$ 0.020 & 0.233 $\pm$ 0.014 \\
450 & 24.81 $\pm$ 1.34 & 0.311 $\pm$ 0.014 & 0.235 $\pm$ 0.012 \\
500 & 24.93 $\pm$ 0.77 & 0.310 $\pm$ 0.009 & 0.235 $\pm$ 0.008 \\
\bottomrule
\end{tabular}
\end{table}

\begin{table}[t]
\centering
\small
\caption{Evaluation of Xhosa with \texttt{xlm-roberta-large-xnli}. Measures are provided of means $\pm$ standard deviation.}
\label{tab:xlmr-xhosa}
\begin{tabular}{rccc}
\toprule
Size & Accuracy (\%) & Precision & F1 \\
\midrule
50  & 29.33 $\pm$ 2.49 & 0.369 $\pm$ 0.043 & 0.316 $\pm$ 0.031 \\
100 & 29.33 $\pm$ 2.87 & 0.346 $\pm$ 0.027 & 0.311 $\pm$ 0.027 \\
150 & 29.78 $\pm$ 3.62 & 0.354 $\pm$ 0.019 & 0.317 $\pm$ 0.031 \\
200 & 30.33 $\pm$ 3.09 & 0.357 $\pm$ 0.017 & 0.324 $\pm$ 0.027 \\
250 & 29.60 $\pm$ 2.36 & 0.354 $\pm$ 0.024 & 0.319 $\pm$ 0.024 \\
300 & 29.33 $\pm$ 2.87 & 0.348 $\pm$ 0.028 & 0.315 $\pm$ 0.029 \\
350 & 29.62 $\pm$ 2.58 & 0.345 $\pm$ 0.027 & 0.316 $\pm$ 0.026 \\
400 & 29.92 $\pm$ 1.12 & 0.349 $\pm$ 0.013 & 0.320 $\pm$ 0.012 \\
450 & 30.00 $\pm$ 0.54 & 0.353 $\pm$ 0.005 & 0.321 $\pm$ 0.003 \\
500 & 29.80 $\pm$ 0.43 & 0.349 $\pm$ 0.007 & 0.319 $\pm$ 0.004 \\
\bottomrule
\end{tabular}
\end{table}

\begin{table}[t]
\centering
\small
\caption{Evaluation of Wolof with \texttt{xlm-roberta-large-xnli}. Measures are provided of means $\pm$ standard deviation.}
\label{tab:xlmr-wolof}
\begin{tabular}{rccc}
\toprule
Size & Accuracy (\%) & Precision & F1 \\
\midrule
50  & 33.33 $\pm$ 5.25 & 0.396 $\pm$ 0.058 & 0.306 $\pm$ 0.024 \\
100 & 32.67 $\pm$ 2.62 & 0.398 $\pm$ 0.040 & 0.311 $\pm$ 0.020 \\
150 & 33.33 $\pm$ 0.54 & 0.392 $\pm$ 0.015 & 0.321 $\pm$ 0.008 \\
200 & 32.67 $\pm$ 1.70 & 0.399 $\pm$ 0.021 & 0.317 $\pm$ 0.018 \\
250 & 31.68 $\pm$ 1.50 & 0.389 $\pm$ 0.030 & 0.305 $\pm$ 0.016 \\
300 & 32.80 $\pm$ 1.66 & 0.375 $\pm$ 0.034 & 0.306 $\pm$ 0.017 \\
350 & 31.24 $\pm$ 1.28 & 0.373 $\pm$ 0.023 & 0.301 $\pm$ 0.013 \\
400 & 31.58 $\pm$ 0.47 & 0.383 $\pm$ 0.008 & 0.304 $\pm$ 0.002 \\
450 & 31.26 $\pm$ 0.38 & 0.379 $\pm$ 0.004 & 0.301 $\pm$ 0.003 \\
500 & 30.53 $\pm$ 0.57 & 0.375 $\pm$ 0.007 & 0.294 $\pm$ 0.006 \\
\bottomrule
\end{tabular}
\end{table}

\begin{table}[t]
\centering
\small
\caption{Evaluation of Luganda with \texttt{xlm-roberta-large-xnli}. Measures are provided of means $\pm$ standard deviation.}
\label{tab:xlmr-luganda}
\begin{tabular}{rccc}
\toprule
Size & Accuracy (\%) & Precision & F1 \\
\midrule
50  & 32.80 $\pm$ 1.63 & 0.470 $\pm$ 0.037 & 0.324 $\pm$ 0.020 \\
100 & 30.08 $\pm$ 0.82 & 0.420 $\pm$ 0.020 & 0.294 $\pm$ 0.004 \\
150 & 32.22 $\pm$ 0.83 & 0.428 $\pm$ 0.023 & 0.315 $\pm$ 0.005 \\
200 & 32.58 $\pm$ 0.71 & 0.421 $\pm$ 0.013 & 0.325 $\pm$ 0.003 \\
250 & 31.68 $\pm$ 0.86 & 0.402 $\pm$ 0.009 & 0.314 $\pm$ 0.009 \\
300 & 32.22 $\pm$ 0.79 & 0.395 $\pm$ 0.007 & 0.318 $\pm$ 0.007 \\
350 & 31.24 $\pm$ 0.49 & 0.384 $\pm$ 0.007 & 0.318 $\pm$ 0.005 \\
400 & 31.25 $\pm$ 0.35 & 0.388 $\pm$ 0.006 & 0.311 $\pm$ 0.002 \\
450 & 31.63 $\pm$ 0.28 & 0.395 $\pm$ 0.003 & 0.315 $\pm$ 0.004 \\
500 & 31.53 $\pm$ 0.34 & 0.393 $\pm$ 0.002 & 0.313 $\pm$ 0.004 \\
\bottomrule
\end{tabular}
\end{table}

\begin{table}[t]
\centering
\small
\caption{Evaluation of Ewe with \texttt{xlm-roberta-large-xnli}. Measures are provided of means $\pm$ standard deviation.}
\label{tab:xlmr-ewe}
\begin{tabular}{rccc}
\toprule
Size & Accuracy (\%) & Precision & F1 \\
\midrule
50  & 36.00 $\pm$ 3.27 & 0.424 $\pm$ 0.154 & 0.318 $\pm$ 0.049 \\
100 & 33.00 $\pm$ 2.16 & 0.468 $\pm$ 0.042 & 0.284 $\pm$ 0.018 \\
150 & 32.22 $\pm$ 1.26 & 0.417 $\pm$ 0.012 & 0.278 $\pm$ 0.010 \\
200 & 30.08 $\pm$ 1.47 & 0.390 $\pm$ 0.010 & 0.260 $\pm$ 0.011 \\
250 & 29.73 $\pm$ 1.86 & 0.385 $\pm$ 0.025 & 0.255 $\pm$ 0.015 \\
300 & 30.22 $\pm$ 1.40 & 0.388 $\pm$ 0.020 & 0.260 $\pm$ 0.011 \\
350 & 29.52 $\pm$ 1.52 & 0.392 $\pm$ 0.027 & 0.259 $\pm$ 0.016 \\
400 & 30.00 $\pm$ 0.54 & 0.400 $\pm$ 0.012 & 0.265 $\pm$ 0.005 \\
450 & 29.85 $\pm$ 0.64 & 0.394 $\pm$ 0.013 & 0.261 $\pm$ 0.004 \\
500 & 29.80 $\pm$ 0.16 & 0.393 $\pm$ 0.009 & 0.257 $\pm$ 0.003 \\
\bottomrule
\end{tabular}
\end{table}

\begin{table}[t]
\centering
\small
\caption{Evaluation of Swahili with \texttt{afro-xlmr-large}. Measures are provided of means $\pm$ standard deviation.}
\label{tab:afro-swahili}
\begin{tabular}{rccc}
\toprule
Size & Accuracy (\%) & Precision & F1 \\
\midrule
50  & 37.33 $\pm$ 0.94 & 0.184 $\pm$ 0.067 & 0.240 $\pm$ 0.057 \\
100 & 35.33 $\pm$ 0.94 & 0.169 $\pm$ 0.052 & 0.223 $\pm$ 0.042 \\
150 & 33.33 $\pm$ 0.94 & 0.210 $\pm$ 0.072 & 0.206 $\pm$ 0.040 \\
200 & 34.67 $\pm$ 0.62 & 0.222 $\pm$ 0.082 & 0.219 $\pm$ 0.044 \\
250 & 33.07 $\pm$ 0.19 & 0.202 $\pm$ 0.067 & 0.204 $\pm$ 0.045 \\
300 & 32.89 $\pm$ 0.68 & 0.210 $\pm$ 0.074 & 0.202 $\pm$ 0.044 \\
350 & 32.19 $\pm$ 0.13 & 0.204 $\pm$ 0.077 & 0.197 $\pm$ 0.045 \\
400 & 32.42 $\pm$ 0.62 & 0.204 $\pm$ 0.077 & 0.198 $\pm$ 0.041 \\
450 & 32.96 $\pm$ 0.55 & 0.207 $\pm$ 0.074 & 0.201 $\pm$ 0.038 \\
500 & 33.00 $\pm$ 0.57 & 0.201 $\pm$ 0.068 & 0.201 $\pm$ 0.038 \\
\bottomrule
\end{tabular}
\end{table}

\begin{table}[t]
\centering
\small
\caption{Evaluation of Lingala with \texttt{afro-xlmr-large}. Measures are provided of means $\pm$ standard deviation.}
\label{tab:afro-lingala}
\begin{tabular}{rccc}
\toprule
Size & Accuracy (\%) & Precision & F1 \\
\midrule
50  & 41.33 $\pm$ 3.40 & 0.246 $\pm$ 0.073 & 0.294 $\pm$ 0.060 \\
100 & 37.67 $\pm$ 1.25 & 0.217 $\pm$ 0.061 & 0.257 $\pm$ 0.050 \\
150 & 37.33 $\pm$ 2.72 & 0.216 $\pm$ 0.072 & 0.253 $\pm$ 0.065 \\
200 & 35.83 $\pm$ 1.43 & 0.202 $\pm$ 0.061 & 0.238 $\pm$ 0.051 \\
250 & 33.87 $\pm$ 0.68 & 0.187 $\pm$ 0.054 & 0.218 $\pm$ 0.039 \\
300 & 33.56 $\pm$ 0.42 & 0.187 $\pm$ 0.053 & 0.215 $\pm$ 0.035 \\
350 & 32.86 $\pm$ 0.62 & 0.183 $\pm$ 0.057 & 0.209 $\pm$ 0.042 \\
400 & 32.42 $\pm$ 0.24 & 0.179 $\pm$ 0.053 & 0.207 $\pm$ 0.038 \\
450 & 32.96 $\pm$ 0.55 & 0.207 $\pm$ 0.074 & 0.201 $\pm$ 0.038 \\
500 & 33.00 $\pm$ 0.57 & 0.201 $\pm$ 0.068 & 0.201 $\pm$ 0.038 \\
\bottomrule
\end{tabular}
\end{table}

\begin{table}[t]
\centering
\small
\caption{Evaluation of Igbo with \texttt{afro-xlmr-large}. Measures are provided of means $\pm$ standard deviation.}
\label{tab:afro-igbo}
\begin{tabular}{rccc}
\toprule
Size & Accuracy (\%) & Precision & F1 \\
\midrule
50  & 36.67 $\pm$ 0.94 & 0.164 $\pm$ 0.039 & 0.215 $\pm$ 0.023 \\
100 & 35.67 $\pm$ 1.25 & 0.157 $\pm$ 0.034 & 0.208 $\pm$ 0.018 \\
150 & 34.67 $\pm$ 0.54 & 0.271 $\pm$ 0.148 & 0.201 $\pm$ 0.031 \\
200 & 34.50 $\pm$ 0.71 & 0.262 $\pm$ 0.137 & 0.201 $\pm$ 0.025 \\
250 & 32.93 $\pm$ 0.19 & 0.254 $\pm$ 0.135 & 0.187 $\pm$ 0.025 \\
300 & 32.89 $\pm$ 0.87 & 0.261 $\pm$ 0.135 & 0.186 $\pm$ 0.029 \\
350 & 32.48 $\pm$ 0.49 & 0.259 $\pm$ 0.139 & 0.184 $\pm$ 0.031 \\
400 & 32.75 $\pm$ 0.41 & 0.216 $\pm$ 0.084 & 0.187 $\pm$ 0.032 \\
450 & 32.96 $\pm$ 0.42 & 0.217 $\pm$ 0.080 & 0.189 $\pm$ 0.027 \\
500 & 33.13 $\pm$ 0.19 & 0.217 $\pm$ 0.082 & 0.189 $\pm$ 0.027 \\
\bottomrule
\end{tabular}
\end{table}

\begin{table}[t]
\centering
\small
\caption{Evaluation of Hausa with \texttt{afro-xlmr-large}. Measures are provided of means $\pm$ standard deviation.}
\label{tab:afro-hausa}
\begin{tabular}{rccc}
\toprule
Size & Accuracy (\%) & Precision & F1 \\
\midrule
50  & 36.67 $\pm$ 0.94 & 0.238 $\pm$ 0.064 & 0.236 $\pm$ 0.030 \\
100 & 35.67 $\pm$ 0.47 & 0.206 $\pm$ 0.054 & 0.223 $\pm$ 0.032 \\
150 & 33.56 $\pm$ 0.31 & 0.192 $\pm$ 0.054 & 0.208 $\pm$ 0.033 \\
200 & 33.67 $\pm$ 0.24 & 0.179 $\pm$ 0.050 & 0.201 $\pm$ 0.034 \\
250 & 32.40 $\pm$ 0.86 & 0.171 $\pm$ 0.051 & 0.188 $\pm$ 0.037 \\
300 & 31.09 $\pm$ 1.50 & 0.163 $\pm$ 0.046 & 0.185 $\pm$ 0.037 \\
350 & 31.43 $\pm$ 0.62 & 0.157 $\pm$ 0.050 & 0.188 $\pm$ 0.037 \\
400 & 31.83 $\pm$ 0.31 & 0.163 $\pm$ 0.048 & 0.185 $\pm$ 0.034 \\
450 & 32.15 $\pm$ 1.00 & 0.169 $\pm$ 0.042 & 0.188 $\pm$ 0.028 \\
500 & 32.27 $\pm$ 0.84 & 0.166 $\pm$ 0.040 & 0.188 $\pm$ 0.028 \\
\bottomrule
\end{tabular}
\end{table}

\begin{table}[t]
\centering
\small
\caption{Evaluation of Zulu with \texttt{afro-xlmr-large}. Measures are provided of means $\pm$ standard deviation.}
\label{tab:afro-zulu}
\begin{tabular}{rccc}
\toprule
Size & Accuracy (\%) & Precision & F1 \\
\midrule
50  & 38.00 $\pm$ 1.63 & 0.148 $\pm$ 0.017 & 0.213 $\pm$ 0.020 \\
100 & 37.00 $\pm$ 0.82 & 0.192 $\pm$ 0.083 & 0.205 $\pm$ 0.015 \\
150 & 35.11 $\pm$ 1.57 & 0.196 $\pm$ 0.114 & 0.189 $\pm$ 0.023 \\
200 & 34.17 $\pm$ 0.24 & 0.174 $\pm$ 0.081 & 0.181 $\pm$ 0.009 \\
250 & 32.88 $\pm$ 0.86 & 0.166 $\pm$ 0.085 & 0.167 $\pm$ 0.012 \\
300 & 32.22 $\pm$ 1.23 & 0.138 $\pm$ 0.047 & 0.162 $\pm$ 0.012 \\
350 & 31.81 $\pm$ 0.49 & 0.136 $\pm$ 0.050 & 0.158 $\pm$ 0.008 \\
400 & 32.33 $\pm$ 0.12 & 0.138 $\pm$ 0.048 & 0.162 $\pm$ 0.005 \\
450 & 32.59 $\pm$ 0.69 & 0.159 $\pm$ 0.058 & 0.165 $\pm$ 0.004 \\
500 & 32.73 $\pm$ 0.50 & 0.149 $\pm$ 0.057 & 0.165 $\pm$ 0.003 \\
\bottomrule
\end{tabular}
\end{table}

\begin{table}[t]
\centering
\small
\caption{Evaluation of Yoruba with \texttt{afro-xlmr-large}. Measures are provided of means $\pm$ standard deviation.}
\label{tab:afro-yoruba}
\begin{tabular}{rccc}
\toprule
Size & Accuracy (\%) & Precision & F1 \\
\midrule
50  & 36.00 $\pm$ 4.32 & 0.220 $\pm$ 0.069 & 0.239 $\pm$ 0.024 \\
100 & 36.33 $\pm$ 1.25 & 0.210 $\pm$ 0.057 & 0.244 $\pm$ 0.041 \\
150 & 34.89 $\pm$ 0.63 & 0.208 $\pm$ 0.066 & 0.228 $\pm$ 0.049 \\
200 & 35.17 $\pm$ 1.03 & 0.201 $\pm$ 0.061 & 0.233 $\pm$ 0.051 \\
250 & 33.87 $\pm$ 0.94 & 0.196 $\pm$ 0.061 & 0.221 $\pm$ 0.050 \\
300 & 33.33 $\pm$ 0.47 & 0.193 $\pm$ 0.056 & 0.216 $\pm$ 0.042 \\
350 & 32.76 $\pm$ 0.71 & 0.190 $\pm$ 0.062 & 0.214 $\pm$ 0.048 \\
400 & 32.92 $\pm$ 0.51 & 0.188 $\pm$ 0.059 & 0.214 $\pm$ 0.046 \\
450 & 33.33 $\pm$ 0.48 & 0.188 $\pm$ 0.053 & 0.217 $\pm$ 0.041 \\
500 & 33.27 $\pm$ 0.34 & 0.191 $\pm$ 0.057 & 0.216 $\pm$ 0.038 \\
\bottomrule
\end{tabular}
\end{table}

\begin{table}[t]
\centering
\small
\caption{Evaluation of Kinyarwanda with \texttt{afro-xlmr-large}. Measures are provided of means $\pm$ standard deviation.}
\label{tab:afro-kinyarwanda}
\begin{tabular}{rccc}
\toprule
Size & Accuracy (\%) & Precision & F1 \\
\midrule
50  & 34.67 $\pm$ 3.40 & 0.166 $\pm$ 0.041 & 0.219 $\pm$ 0.027 \\
100 & 36.00 $\pm$ 1.63 & 0.280 $\pm$ 0.141 & 0.231 $\pm$ 0.038 \\
150 & 33.11 $\pm$ 1.26 & 0.189 $\pm$ 0.052 & 0.205 $\pm$ 0.038 \\
200 & 33.17 $\pm$ 1.55 & 0.185 $\pm$ 0.049 & 0.206 $\pm$ 0.033 \\
250 & 32.67 $\pm$ 0.50 & 0.182 $\pm$ 0.051 & 0.200 $\pm$ 0.046 \\
300 & 32.44 $\pm$ 1.50 & 0.172 $\pm$ 0.045 & 0.197 $\pm$ 0.049 \\
350 & 31.71 $\pm$ 0.62 & 0.168 $\pm$ 0.048 & 0.191 $\pm$ 0.049 \\
400 & 32.17 $\pm$ 0.12 & 0.165 $\pm$ 0.047 & 0.194 $\pm$ 0.046 \\
450 & 32.52 $\pm$ 0.73 & 0.166 $\pm$ 0.041 & 0.196 $\pm$ 0.041 \\
500 & 32.33 $\pm$ 0.82 & 0.162 $\pm$ 0.039 & 0.194 $\pm$ 0.038 \\
\bottomrule
\end{tabular}
\end{table}

\begin{table}[t]
\centering
\small
\caption{Evaluation of Amharic with \texttt{afro-xlmr-large}. Measures are provided of means $\pm$ standard deviation.}
\label{tab:afro-amharic}
\begin{tabular}{rccc}
\toprule
Size & Accuracy (\%) & Precision & F1 \\
\midrule
50  & 36.67 $\pm$ 0.94 & 0.167 $\pm$ 0.043 & 0.223 $\pm$ 0.033 \\
100 & 37.33 $\pm$ 1.89 & 0.186 $\pm$ 0.076 & 0.229 $\pm$ 0.050 \\
150 & 34.67 $\pm$ 1.44 & 0.163 $\pm$ 0.068 & 0.203 $\pm$ 0.045 \\
200 & 34.33 $\pm$ 0.85 & 0.161 $\pm$ 0.064 & 0.208 $\pm$ 0.038 \\
250 & 32.80 $\pm$ 1.18 & 0.149 $\pm$ 0.062 & 0.183 $\pm$ 0.035 \\
300 & 32.67 $\pm$ 1.66 & 0.152 $\pm$ 0.069 & 0.183 $\pm$ 0.038 \\
350 & 32.57 $\pm$ 1.68 & 0.181 $\pm$ 0.064 & 0.183 $\pm$ 0.041 \\
400 & 33.08 $\pm$ 1.24 & 0.192 $\pm$ 0.065 & 0.188 $\pm$ 0.037 \\
450 & 33.04 $\pm$ 0.73 & 0.184 $\pm$ 0.053 & 0.188 $\pm$ 0.029 \\
500 & 33.07 $\pm$ 0.47 & 0.182 $\pm$ 0.052 & 0.187 $\pm$ 0.027 \\
\bottomrule
\end{tabular}
\end{table}

\begin{table}[t]
\centering
\small
\caption{Evaluation of Southern sotho with \texttt{afro-xlmr-large}. Measures are provided of means $\pm$ standard deviation.}
\label{tab:afro-southern-sotho}
\begin{tabular}{rccc}
\toprule
Size & Accuracy (\%) & Precision & F1 \\
\midrule
50  & 39.33 $\pm$ 3.40 & 0.202 $\pm$ 0.092 & 0.238 $\pm$ 0.054 \\
100 & 37.08 $\pm$ 0.82 & 0.174 $\pm$ 0.058 & 0.211 $\pm$ 0.022 \\
150 & 34.67 $\pm$ 0.54 & 0.264 $\pm$ 0.150 & 0.191 $\pm$ 0.017 \\
200 & 34.50 $\pm$ 0.71 & 0.261 $\pm$ 0.137 & 0.192 $\pm$ 0.014 \\
250 & 33.33 $\pm$ 0.50 & 0.226 $\pm$ 0.089 & 0.182 $\pm$ 0.018 \\
300 & 33.08 $\pm$ 0.94 & 0.226 $\pm$ 0.087 & 0.179 $\pm$ 0.019 \\
350 & 32.57 $\pm$ 0.81 & 0.208 $\pm$ 0.075 & 0.174 $\pm$ 0.022 \\
400 & 32.67 $\pm$ 0.42 & 0.184 $\pm$ 0.058 & 0.177 $\pm$ 0.018 \\
450 & 32.89 $\pm$ 0.48 & 0.193 $\pm$ 0.057 & 0.179 $\pm$ 0.009 \\
500 & 33.07 $\pm$ 0.25 & 0.192 $\pm$ 0.057 & 0.180 $\pm$ 0.011 \\
\bottomrule
\end{tabular}
\end{table}

\begin{table}[t]
\centering
\small
\caption{Evaluation of Oromo with \texttt{afro-xlmr-large}. Measures are provided of means $\pm$ standard deviation.}
\label{tab:afro-oromo}
\begin{tabular}{rccc}
\toprule
Size & Accuracy (\%) & Precision & F1 \\
\midrule
50  & 38.00 $\pm$ 3.27 & 0.186 $\pm$ 0.038 & 0.230 $\pm$ 0.016 \\
100 & 37.33 $\pm$ 1.25 & 0.287 $\pm$ 0.143 & 0.228 $\pm$ 0.030 \\
150 & 35.56 $\pm$ 1.13 & 0.231 $\pm$ 0.082 & 0.212 $\pm$ 0.034 \\
200 & 34.83 $\pm$ 0.62 & 0.216 $\pm$ 0.073 & 0.204 $\pm$ 0.030 \\
250 & 33.33 $\pm$ 0.50 & 0.204 $\pm$ 0.073 & 0.190 $\pm$ 0.022 \\
300 & 33.08 $\pm$ 0.54 & 0.186 $\pm$ 0.052 & 0.187 $\pm$ 0.021 \\
350 & 32.19 $\pm$ 0.49 & 0.177 $\pm$ 0.053 & 0.179 $\pm$ 0.023 \\
400 & 32.50 $\pm$ 0.35 & 0.179 $\pm$ 0.053 & 0.182 $\pm$ 0.025 \\
450 & 32.81 $\pm$ 0.52 & 0.192 $\pm$ 0.058 & 0.185 $\pm$ 0.023 \\
500 & 33.00 $\pm$ 0.33 & 0.193 $\pm$ 0.059 & 0.186 $\pm$ 0.025 \\
\bottomrule
\end{tabular}
\end{table}

\begin{table}[t]
\centering
\small
\caption{Evaluation of Twi with \texttt{afro-xlmr-large}. Measures are provided of means $\pm$ standard deviation.}
\label{tab:afro-twi}
\begin{tabular}{rccc}
\toprule
Size & Accuracy (\%) & Precision & F1 \\
\midrule
50  & 37.33 $\pm$ 0.94 & 0.179 $\pm$ 0.058 & 0.236 $\pm$ 0.049 \\
100 & 35.33 $\pm$ 1.25 & 0.194 $\pm$ 0.048 & 0.238 $\pm$ 0.034 \\
150 & 33.11 $\pm$ 0.83 & 0.188 $\pm$ 0.053 & 0.213 $\pm$ 0.033 \\
200 & 33.33 $\pm$ 1.03 & 0.186 $\pm$ 0.051 & 0.213 $\pm$ 0.032 \\
250 & 31.73 $\pm$ 1.00 & 0.175 $\pm$ 0.047 & 0.199 $\pm$ 0.030 \\
300 & 31.67 $\pm$ 1.25 & 0.177 $\pm$ 0.046 & 0.197 $\pm$ 0.028 \\
350 & 31.05 $\pm$ 0.75 & 0.171 $\pm$ 0.049 & 0.198 $\pm$ 0.031 \\
400 & 31.83 $\pm$ 0.62 & 0.177 $\pm$ 0.052 & 0.196 $\pm$ 0.033 \\
450 & 31.85 $\pm$ 1.21 & 0.173 $\pm$ 0.044 & 0.195 $\pm$ 0.025 \\
500 & 32.13 $\pm$ 1.09 & 0.175 $\pm$ 0.046 & 0.196 $\pm$ 0.026 \\
\bottomrule
\end{tabular}
\end{table}

\begin{table}[t]
\centering
\small
\caption{Evaluation of Shona with \texttt{afro-xlmr-large}. Measures are provided of means $\pm$ standard deviation.}
\label{tab:afro-shona}
\begin{tabular}{rccc}
\toprule
Size & Accuracy (\%) & Precision & F1 \\
\midrule
50  & 36.00 $\pm$ 1.63 & 0.174 $\pm$ 0.053 & 0.229 $\pm$ 0.042 \\
100 & 36.00 $\pm$ 0.00 & 0.173 $\pm$ 0.058 & 0.228 $\pm$ 0.050 \\
150 & 33.78 $\pm$ 0.31 & 0.156 $\pm$ 0.058 & 0.208 $\pm$ 0.051 \\
200 & 34.00 $\pm$ 0.00 & 0.155 $\pm$ 0.055 & 0.208 $\pm$ 0.048 \\
250 & 32.67 $\pm$ 0.75 & 0.146 $\pm$ 0.057 & 0.196 $\pm$ 0.052 \\
300 & 31.67 $\pm$ 1.44 & 0.139 $\pm$ 0.049 & 0.187 $\pm$ 0.044 \\
350 & 31.52 $\pm$ 0.49 & 0.138 $\pm$ 0.054 & 0.186 $\pm$ 0.049 \\
400 & 32.50 $\pm$ 0.54 & 0.144 $\pm$ 0.057 & 0.194 $\pm$ 0.052 \\
450 & 32.67 $\pm$ 0.65 & 0.144 $\pm$ 0.051 & 0.195 $\pm$ 0.045 \\
500 & 32.53 $\pm$ 0.74 & 0.180 $\pm$ 0.048 & 0.194 $\pm$ 0.049 \\
\bottomrule
\end{tabular}
\end{table}

\begin{table}[t]
\centering
\small
\caption{Evaluation of Xhosa with \texttt{afro-xlmr-large}. Measures are provided of means $\pm$ standard deviation.}
\label{tab:afro-xhosa}
\begin{tabular}{rccc}
\toprule
Size & Accuracy (\%) & Precision & F1 \\
\midrule
50  & 39.33 $\pm$ 2.49 & 0.258 $\pm$ 0.149 & 0.238 $\pm$ 0.028 \\
100 & 37.67 $\pm$ 0.94 & 0.358 $\pm$ 0.156 & 0.214 $\pm$ 0.013 \\
150 & 35.33 $\pm$ 1.44 & 0.347 $\pm$ 0.164 & 0.198 $\pm$ 0.016 \\
200 & 35.00 $\pm$ 0.41 & 0.306 $\pm$ 0.193 & 0.189 $\pm$ 0.009 \\
250 & 33.47 $\pm$ 1.05 & 0.307 $\pm$ 0.153 & 0.174 $\pm$ 0.014 \\
300 & 32.89 $\pm$ 1.40 & 0.277 $\pm$ 0.146 & 0.169 $\pm$ 0.016 \\
350 & 32.38 $\pm$ 0.71 & 0.260 $\pm$ 0.137 & 0.163 $\pm$ 0.018 \\
400 & 32.83 $\pm$ 0.31 & 0.263 $\pm$ 0.135 & 0.167 $\pm$ 0.006 \\
450 & 32.89 $\pm$ 0.63 & 0.219 $\pm$ 0.093 & 0.167 $\pm$ 0.007 \\
500 & 33.00 $\pm$ 0.43 & 0.219 $\pm$ 0.092 & 0.168 $\pm$ 0.005 \\
\bottomrule
\end{tabular}
\end{table}

\begin{table}[t]
\centering
\small
\caption{Evaluation of Wolof with \texttt{afro-xlmr-large}. Measures are provided of means $\pm$ standard deviation.}
\label{tab:afro-wolof}
\begin{tabular}{rccc}
\toprule
Size & Accuracy (\%) & Precision & F1 \\
\midrule
50  & 39.33 $\pm$ 1.89 & 0.236 $\pm$ 0.067 & 0.241 $\pm$ 0.031 \\
100 & 38.08 $\pm$ 1.63 & 0.223 $\pm$ 0.076 & 0.226 $\pm$ 0.030 \\
150 & 34.67 $\pm$ 1.44 & 0.172 $\pm$ 0.055 & 0.198 $\pm$ 0.028 \\
200 & 34.17 $\pm$ 0.24 & 0.171 $\pm$ 0.040 & 0.194 $\pm$ 0.015 \\
250 & 32.67 $\pm$ 0.50 & 0.166 $\pm$ 0.041 & 0.183 $\pm$ 0.013 \\
300 & 32.11 $\pm$ 1.10 & 0.159 $\pm$ 0.034 & 0.177 $\pm$ 0.006 \\
350 & 31.81 $\pm$ 0.13 & 0.160 $\pm$ 0.041 & 0.174 $\pm$ 0.014 \\
400 & 32.17 $\pm$ 0.31 & 0.160 $\pm$ 0.040 & 0.177 $\pm$ 0.015 \\
450 & 32.44 $\pm$ 1.01 & 0.163 $\pm$ 0.040 & 0.179 $\pm$ 0.013 \\
500 & 32.73 $\pm$ 0.81 & 0.170 $\pm$ 0.044 & 0.182 $\pm$ 0.015 \\
\bottomrule
\end{tabular}
\end{table}

\begin{table}[t]
\centering
\small
\caption{Evaluation of Luganda with \texttt{afro-xlmr-large}. Measures are provided of means $\pm$ standard deviation.}
\label{tab:afro-luganda}
\begin{tabular}{rccc}
\toprule
Size & Accuracy (\%) & Precision & F1 \\
\midrule
50  & 41.33 $\pm$ 3.40 & 0.298 $\pm$ 0.140 & 0.282 $\pm$ 0.052 \\
100 & 35.67 $\pm$ 2.05 & 0.219 $\pm$ 0.072 & 0.235 $\pm$ 0.031 \\
150 & 34.22 $\pm$ 2.45 & 0.210 $\pm$ 0.075 & 0.222 $\pm$ 0.035 \\
200 & 33.33 $\pm$ 1.31 & 0.199 $\pm$ 0.065 & 0.213 $\pm$ 0.030 \\
250 & 32.67 $\pm$ 1.05 & 0.196 $\pm$ 0.067 & 0.205 $\pm$ 0.030 \\
300 & 32.08 $\pm$ 1.36 & 0.184 $\pm$ 0.052 & 0.198 $\pm$ 0.026 \\
350 & 31.43 $\pm$ 0.62 & 0.176 $\pm$ 0.053 & 0.192 $\pm$ 0.032 \\
400 & 31.67 $\pm$ 0.82 & 0.176 $\pm$ 0.052 & 0.194 $\pm$ 0.031 \\
450 & 32.44 $\pm$ 0.79 & 0.181 $\pm$ 0.049 & 0.201 $\pm$ 0.033 \\
500 & 32.93 $\pm$ 0.66 & 0.183 $\pm$ 0.051 & 0.205 $\pm$ 0.038 \\
\bottomrule
\end{tabular}
\end{table}

\begin{table}[t]
\centering
\small
\caption{Evaluation of Ewe with \texttt{afro-xlmr-large}. Measures are provided of means $\pm$ standard deviation.}
\label{tab:afro-ewe}
\begin{tabular}{rccc}
\toprule
Size & Accuracy (\%) & Precision & F1 \\
\midrule
50  & 39.33 $\pm$ 1.89 & 0.233 $\pm$ 0.070 & 0.260 $\pm$ 0.042 \\
100 & 38.00 $\pm$ 2.83 & 0.226 $\pm$ 0.071 & 0.252 $\pm$ 0.051 \\
150 & 36.44 $\pm$ 1.75 & 0.224 $\pm$ 0.078 & 0.241 $\pm$ 0.050 \\
200 & 35.33 $\pm$ 1.25 & 0.203 $\pm$ 0.063 & 0.227 $\pm$ 0.041 \\
250 & 33.20 $\pm$ 0.33 & 0.188 $\pm$ 0.055 & 0.207 $\pm$ 0.029 \\
300 & 32.78 $\pm$ 0.83 & 0.185 $\pm$ 0.051 & 0.208 $\pm$ 0.022 \\
350 & 32.67 $\pm$ 0.75 & 0.192 $\pm$ 0.064 & 0.208 $\pm$ 0.031 \\
400 & 33.08 $\pm$ 0.59 & 0.192 $\pm$ 0.063 & 0.203 $\pm$ 0.033 \\
450 & 33.26 $\pm$ 0.58 & 0.193 $\pm$ 0.057 & 0.204 $\pm$ 0.027 \\
500 & 33.53 $\pm$ 0.82 & 0.195 $\pm$ 0.059 & 0.207 $\pm$ 0.030 \\
\bottomrule
\end{tabular}
\end{table}

\end{document}